# DEVELOPMENT OF CROP YIELD ESTIMATION MODEL USING SOIL AND ENVIRONMENTAL PARAMETERS

*Nisar Ahmed [1], Hafiz Muhammad Shahzad Asif [2], Gulshan Saleem [3] and Muhammad Usman Younus * [4]*

[1] Ph.D. Scholar, [2] Associate Professor, Department of Computer Science and Engineering, University of Engineering and Technology, Lahore, [3] Ph.D. Scholar, Department of Computer Science, COMSATS University Islamabad (Lahore Campus), Lahore, [4] Researcher, Ecole Doctorale Matheematiques, Informatique, Telecommunication de Toulouse, Paul Sabatier University, Toulouse, France

*Corresponding author's email: usman1644@gmail.com

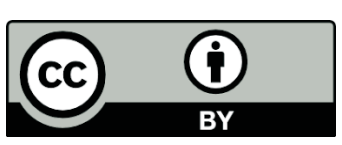

## ABSTRACT

Current study was conducted in collaboration with Department of Computer Sciences and Engineering, University of Engineering and technology, Lahore, Pakistan during 2020. As a result, the focus of this work was on the development of a pre-harvest crop yield forecasting model based on soil and environmental parameters. These parameters were recorded on a monthly basis at National Tea & High Value Crops Research Institute (NTHRI) for a period of ten years. The parameters recorded were minimum and maximum temperature, humidity, rainfall, soil pH level, pesticide use, and labor expertise. To construct the feature set for model training, exploratory feature analysis, outlier analysis, feature scaling, feature transformation and feature selection using the ReliefF algorithm were performed. Through 10-fold cross validation, six regression algorithms were used for training and model evaluation. An ensemble of neural networks were used to build the final model. The ensemble were built using a novel method that trains several base learners with architectural and training data diversity. These trained models were ranked using the ReliefF algorithm and sequentially added to the ensemble until the validation performance stops improving. On the basis of four performance metrics, the final model was compared to the four best performing models. The proposed model provided mean averave error (MAE), mean squared error (MSE) and root mean squared error (RMSE) of 0.0942, 0.0145, and 0.1204, respectively, and an R-squared of 0.9461. The performance parameters were the best among the candidate models and were sufficient to justify its use as a tea crop yield prediction model.

## INTRODUCTION

Farming has a major role in the growth of human civilization as it is the only way to support the increasing food demands. With the growth of human population and settlements, the farmland area decreased and the need of food supplies increased. Advancements in agricultural technologies proven helpful to cater for increased needs and reduced crop loss. Plant genetics, weed and pest control methodologies and fertilizers are employed to overcome the decreasing soil fertility and increase the yield. Weather and soil analysis has provided insight to the parameters affecting the growth and yield and therefore are helpful in identification of suitable crop for a particular weather and land type. Field monitoring is important as it is helpful in yield estimation and contributes towards economic development and food security of a nation. Remote sensing has proven to be effective in agricultural monitoring, yield forecasting and irrigation scheduling (Weiss *et al.,* 2020; Ahmad *et al.,* 2021). These techniques provide flood and other disaster warnings as well as crop growth monitoring and disease spread modeling. The crop production and yield prediction have direct influence on the economics and food management (Hayes and Decker, 1996; Kogan *et al.,* 2019)). Estimation of crop yield is of particular interest to various stakeholders: farmers, national government and international institutions to strengthen the food security. The crop yield is considered as indicator of productivity which is total harvested mass per unit area. Small farm holders typically focus on yield at farm scale, expecting to increase their financial returns. Whereas government, international organizations and commodity traders are focused to yield estimates at larger scales such as administrative units or national levels. Crop insurer focus towards both small scale and large-scale yield estimates in order to assess field loss at finer scale and aggregate the yield at regional or larger scale. These type of information are relevant to policy-makers to provide guidelines for market intervention, trade and assistance (Filippi *et al.,* 2019; Kogan *et al.,* 2019).

Advances in Information Technology (IT) has provided the way to agricultural management. Processing of remotely sensed multispectral data, calculation of different reflectance indices, crop cover identification, disease identification & spread modeling, weather monitoring and forecasting, decision support systems and yield estimation are some of the techniques which are based on IT (Weiss *et al.,* 2020; Younus *et al.,* 2016). Crop yield estimation is usually performed using various methods and data sources such as expert knowledge, field surveys, statistical modeling and crop growth simulation models (Weiss *et al.,* 2020). Identification of crop condition and its potential outcomes can provide useful insights which can be helpful to cater the yield loss (Dutta, 2006). Various crop growth monitoring and yield estimation models are developed by research community which are based on soil, weather and remotely sensed data (Matthews and Stephens, 1998; Matthews *et al.,* 2002; Younus and Kim*,* 2019). Pre-harvest yield estimation therefore plays an important role in maintenance of national food security (Hayes and Decker, 1996).

Tea (*Camellia sinensis*) is a high value crop and among biggest commodity of Pakistan. Pakistan spends large revenue on import of tea and any development in its yield improvement may be beneficial for economy (Ahmed *et al.,* 2021). We have started our work in collaboration with National Tea and High Value Crops Research Institute (NTHRI), Shinkiyari, Mansehra. It is an evergreen shrub and its leaves are plucked to produce green tea, black tea or oolong tea (depends on processing). Its tree is cut to below two meters when cultivated for tea production. Depending on their chemical footprint, different leaf ages produce varying qualities of tea. Typically, small, fresh leaves are plucked for tea every one or two weeks. Tea was originated in Southeast Asia and is cultivated in warm-wet summers and cold-dry winters. The tea plants require annual rainfall with the range of 1150 to 1400 but its distribution throughout the year is equally important. Humidity also has a good relationship with tea yield and the quality of produced tea. A relative humidity of above 90%and temperature range of 20-26°C is very helpful for high yield and good quality. Soil PH is a very good indicator of yield prediction as it is affected with land fertility, land humidity and plant growth because temperature, rainfall and humidity are directly related to this parameter. Temperature at the other had is a very critical parameter as the significantly low or high temperature may entirely stop the growth regardless of other parameters. In the case of Tea farms at Shinkiyari, temperature usually remains in favorable range and doesn't cause catastrophic effect on the growth. Sunshine, fast wind and hail affects the growth as well but in the current data, we don't have access to this information. In his classical review stated that the ideal growing weather for tea is warmer temperature with high humidity and sunshine in the day and adequate rainfall during night time (Laycock, 1964).

In this study, parametric analysis is performed to identify the important parameters and irrelevant or redundant parameters are eliminated. Feature Selection is performed using a hybrid approach which perform feature ranking with ReliefF algorithm and sequentially add these ranked features until model performance stops increasing. These selected and complete feature sets are used to train on different regression algorithms to set a baseline performance as the dataset under use is not previously evaluated by researchers. The final model is constructed by training proposed neural network ensemble on the processed feature set.

## MATERIALS AND METHODS

Current study was conducted in collaboration with Department of Computer Sciences and Engineering, University of Engineering and Technology, Lahore, Pakistan during 2020. The data used in the study was gathered from the farms in Shinkiyari, district Mansehra (latitude, 34°27' E; longitude, 73°16' N; altitude, 542m) which are operated by National Tea & High value crops Research Institute (NTHRI) of Pakistan. The data collected on monthly basis up to ten years, a total of 120 samples. The data contains nine independent variables namely minimum and maximum temperature of the farm, humidity level, rainfall, pH value of soil, per labor cost, labor training level, pesticide use, month and yield. Labor cost were not used in model design as it had no direct effect on yield and its values increased with time. Labor training levels had only two different levels and had shown no correlation with the yield. Similarly, pesticide use was also excluded from model design as it only mention that pesticide was used and there was no mention of its amount or type. Minimum and maximum temperatures were extracted from daily temperature readings recorded in the farms and their maximum and minimum values within a month were recorded. Humidity was the level of humidity recorded within the farms on daily basis and their average value during a month was recorded. Rainfall was the total amount of rain during a month. The pH level was the average value of pH which is measured at a depth of 0.3 m at various sites in the tea farm. Month refers to the categorical variable corresponding to each month of the year and it was found to have good correlation with the yield. Yield was recorded value of crop produce in kilograms for black tea.

Uncertainty in the recorded parameters arose due to spatial variability of weather and soil parameters i.e. soil pH and temperature are taken at some fixed points and their values was usually averaged for a month's period. Average temperature was calculated for each month from minimum and maximum temperature value as it better represent the yield and incorporated as another variable.

However, Table 1 provided the values of Pearson's correlation coefficient test between different parameters. These tests measured the linear relationship of a variable with others. It was notable that average, maximum and minimum temperatures had very high correlation with each other. The reason for this relation was that three of these temperature values varied in a linear relationship to each other's. Some variables had negative linear relationship with each other and indicate that increase in one, might result in decrease in the other variable such as rainfall will result in decreased pH level.

Furthermore, Table 2 provides the values of pearson correlation between the independent variables and yield. It can be noted that all the independent variables have positive correlation with yield except soil pH. Moreover, rainfall had the highest value of correlation coefficient with yield. It could be noted that pH value was negatively correlated with rainfall and have highest inter variable correlation. As a final feature set we have eight parameters (features) as an independent variables and one parameter, yield, as dependent parameter. Crop yield forecasting was done by fitting regression algorithm on the available data and testing on unseen samples. We had tried nine different regression algorithms on the available data and performed cross-validation to find the most suitable one for our scenario. Following are the steps followed to construct the crop yield forecasting model.

### Outlier analysis

Outliers were the samples which had strong influence on the model's performance. These points might be erroneous measurements and caused the model to behave improperly. Outlier analysis was performed to identify the influence points and check them for their validity. Identified outliers are sometimes removed or they were replaced through smoothing. We had used Cook's distance to identify the influence points. We had selected those points with cook's distance larger than 0.5 to treat them as outliers and they were provided in Table 3. Sample ID indicate the index of the outlier sample, yield was the value of yield for the outlier and the last one was cook's distance which was the basis of their selection. The performance of the model before and after the removal of outliers was provided in results section.

**Table 3. Removed outliers: having large cook's distance**

| Tuple ID | Yield | Cook's distance |
|---|---|---|
| 84 | 115.8345 | 0.76 |
| 56 | 44.7655 | 0.58 |
| 60 | 71.6392 | 0.50 |

### Feature scaling

Feature scaling was a fundamental step for a lot of machine learning problems due to lot of reasons. It involved scaling the range of features to a specified range as different features were measured in different units and might had large variation in their range. The need of feature scaling arose as different ranges of features might caused the objective function to work improperly. It is especially important for distance based algorithms as when the features have different ranges, distance measure will be governed by the feature with large range. It was therefore, preferred to normalize the range of all the features so that each feature contribute almost proportionately towards distance measure. In gradient descent and stochastic gradient descent, the convergence speed is sometimes improved after application of feature normalization (Grus, 2015; Loffe and Szegedy, 2015). In support vector machines (SVM), feature scaling can reduce the time to find support vectors and affect the results as well (Juszczak *et al.,* 2002).There were different approaches to feature scaling such as min-max scaling, mean normalization, scaling to unit length and standardization. We had used standardization due to its advantages in neural networks, SVM, perceptron and radial basis function (Grus, 2015). The standardization is performed so that data has zero-mean and unit variance and is calculated with the below formula:

$$x' = \frac{x - x}{\sigma}$$

**Table 1. Inter-parameter pearson correlation coefficient test**

| Parameters | Average temperature | Maximum temperature | Minimum temperature | Humidity | Rainfall | pH scale |
|---|---|---|---|---|---|---|
| Average temperature | 1.000 | 0.967 | 0.969 | 0.132 | 0.248 | 0.048 |
| Maximum temperature | 0.967 | 1.000 | 0.875 | -0.010 | 0.119 | 0.147 |
| Minimum temperature | 0.969 | 0.875 | 1.000 | 0.261 | 0.358 | -0.052 |
| Humidity | 0.132 | -0.010 | 0.261 | 1.000 | 0.354 | -0.241 |
| Rainfall | 0.248 | 0.119 | 0.358 | 0.354 | 1.000 | -0.630 |
| pH scale | 0.048 | 0.147 | -0.052 | -0.241 | -0.630 | 1.000 |

**Table 2. Linear correlation between dependent variables and independent variable**

| Parameter | Average temperature | Maximum temperature | Minimum temperature | Rainfall | Humidity | Soil pH |
|---|---|---|---|---|---|---|
| Correlation with yield | 0.35 | 0.22 | 0.47 | 0.87 | 0.38 | -0.74 |

## Feature selection

There were numerous approaches to features selection with their advantages and limitations. We have opted a hybrid approach which use ReliefF (Robnik-Šikonja and Kononenko, 2003) and sequential feature selection (Aha and Bankert, 1996). ReliefF algorithm works by penalizing the features which gave different values to the neighbors of the same class and reward the features which give different values to the neighbors of the different classes. The algorithm provides the rank of each feature and weight associated to each one of them. Normally the features with positive weights were the ones which provide good prediction and the features with negative weights were the ones which negatively affect the prediction. Fig. 1 provides the ranked features on x-axes and their weight value on the y-axes.

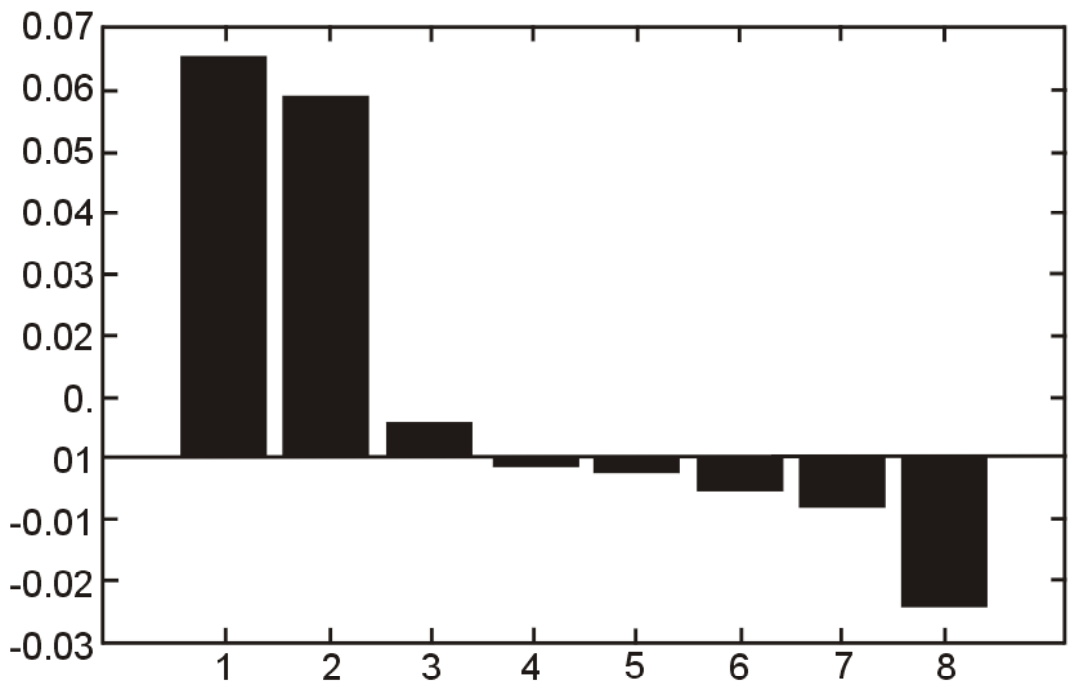

**Fig. 1. Ranked features with their weights calculated through relief**

It was noted from the plot that only three features had positive values and the rest of the features had negative values. We applied forward feature selection approach by sequentially adding each features in the ranked order and cross-validating it for reduction in root mean squared error (RMSE).

We selected linear SVM for cross-validation evaluation of features and selected five features. A RMSE value was achieved using linear SVM with five features as provided in Table 4.

## Feature transformation

Feature transformation sometime helps in improvement of model performance. One purpose of using feature transformation is to compress the dynamic range of data which is useful in many algorithms. We have tried square root, reciprocal and log transformation on the data and checked the cross-validation performance. The log-transformation came out to be the winner of the three and have major effect on the distribution shape of the data. It is demonstrated to reduce the skewness and is appropriate for measured variables. It can't be applied for zero or negative values.

## Baseline performance

Design of any prediction system requires an algorithm which is trained and validated on the dataset. Selection of a suitable regression algorithm is therefore fundamental to efficient model construction. Moreover, the dataset under use was not used for yield forecasting in past and need a baseline performance. To establish a baseline performance, we trained seven different regression algorithms 10-fold Cross-Validation split. The six algorithms used for baseline performance assessment were multiple linear regression (MLR), Support Vector Machines Regression (SVR) with linear and quadratic kernels, Gaussian Process Regression (GPR), Artificial Neural Networks (ANN), random forest (RF) and AdaBoost. The performance of these algorithms were computed on same CV split and performed on raw dataset and processed features for comparison.

**Table 4. Root mean squared error (RMSE) of five regression models after each stage**

| Parameters | Raw dataset | Feature selection | Feature scaling | Outliers removal | Feature transformation |
|---|---|---|---|---|---|
| **GPR** | 7.1456 | 6.6265 | 6.3128 | 5.9961 | 0.1542 |
| **Linear SVR** | 8.1314 | 7.9937 | 7.8867 | 7.778 | 0.2215 |
| **Quadratic SVR** | 7.4107 | 7.3067 | 7.1077 | 6.4612 | 0.1696 |
| **MLR** | 8.2140 | 7.9148 | 7.4444 | 7.4328 | 0.2151 |
| **ANN** | 7.1954 | 6.8541 | 6.2724 | 5.9614 | 0.1674 |
| **RF** | 7.1871 | 6.9821 | 6.4215 | 5.9712 | 0.1705 |
| **AdaBoost** | 7.4831 | 7.1421 | 6.8725 | 6.2143 | 0.2024 |

GPR = Gaussian process regression, MLR = Multiple linear regression, RMS = Root mean squared error, SUR = Support vector regression, ANN = Artifical neural network, RF = Random forest

### Neural networks ensemble

The baseline results indicated that the GPR has minimum root mean squared euror (RMSE) RMSE after hyperparameter optimization. We on the other hand want to opt neural networks ensemble due to some of its interesting properties. Ensemble learning is a very appealing topic due to its incredible simplicity and complex underlying dynamics. Neural networks are good at learning complex relationships between independent and target variables but suffer from instability. Multiple training instances of a single neural networks will provide different validation accuracy due to different convergence point of neural networks. An averaging ensemble can be simply used to average the prediction of multiple trained neural networks which will provide stable results and having good generalization. However, we can do more to improve the performance of neural networks ensemble.

To train a neural network ensemble we constructed a shallow neural network architecture with only one hidden layer having five to thirty neurons in the hidden layer. The number of neurons in the hidden layer were randomly selected for each base learner. The training is performed by providing random subsampling of training data. It was to be noted that validation data was never used in any of the model construction phase, it was only used for validation after the end of ensemble learning process. The training process has trained 100 base learners with random subsampling and varying architectures to introduce diversity in the models. Moreover, the trained base learners were combined on the basis of model selection approach to form ensemble.

### Model selection

It was better to combine few accurate base learners rather than combining all the base learners. Dietterich (2000). has demonstrated that the base learners used for ensemble construction should be diverse and relatively accurate to make it effective. However, it was unlikely for both requirements to be true as if they were accurate (Christensen, 2003). We initially trained 100 base learners having architectural and training diversity and then few accurate one may added in the final ensemble. Wrapper based methods was to add models in the final ensemble by sequentially adding ensemble and cross-validating on a portion of dataset but we opted a hybrid approach for model selection. The base models was ranked using ReliefF algorithm which provided feature weights and their rank. Fig. 2 showed a bar chart of ranked base learners with their weights. The negative weights were assigned to the predictors conversely affecting the output. We started sequentially adding base learners into the final model until there was no increase in the performance. Twenty-four base learners are sequentially added into final model until there was increase in the performance and further addition of base learners provided no change in predictive performance.

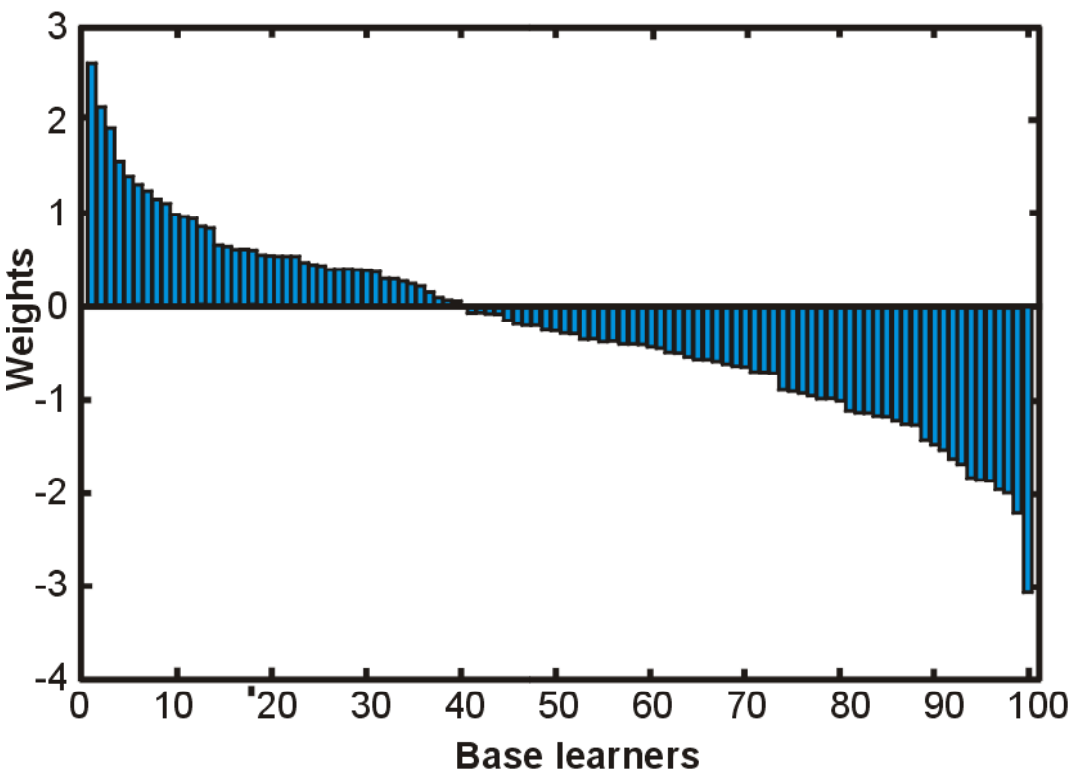

**Fig. 2. Bar chart of ranked base learners with their assigned weights**

### Output calculation

The twenty-four base learners sequentially obtained in model selection process were used for output calculation. We had calculated the output function by initially assigning higher weight to the learners with low mean squared error $\epsilon_i$. Weights were calculated for these predictors using a convex function to assign lower weight to a neural network with higher error and higher weight to a neural network with lower error. The weight $\omega_i$ of $ith$ weak learner was calculated using formula as under:

$$\omega_i = \frac{1}{e^{-b(|\epsilon_i| - c)}}$$

The final output of predicted value was calculated by weighted average of 24 base learners which was given by the formula below:-

$$Y_{pred} = \sum_{i=1}^{10} \omega_i \times Y_i$$

The flow chart of the proposed ensemble learning process is provided in Fig. 3.

## RESULTS AND DISCUSSION

Data (Table 1) provided the values of independent variables and target variables. It was evident that rainfall demonstrates best linear relationship. pH also had somewhat linear relationship but temperature and humidity parameters had poor linear relation. pH was observed to be highly impactful variable in determining the tea yield followed by rainfall, month of the year, temperature and humidity. Moreover, pH scale had the highest correlation with tea yield. It was indicated that pH scale had negative relationship with yield whereas the other parameters such as rainfall, average temperature, humidity had positive linear relationship based on calculated coefficients. It was proved to be difficult to map the data by using linear model but we adopted a modular approach and found the cross-validation accuracy at each stage.

In first stage we discarded the features which had no relationship with the target variables based on empirical analysis such as labor cost, labor training, pesticide and year of harvest as some of them have consistent values. After that feature selection was performed by first ranking them with ReliefF algorithm and then applying sequential feature selection to add each ranked feature one by one to get the best feature set. The selected feature set contained five features. Table 4 provided the improvement in the results after feature selection stage.

The next stages were feature selection and outlier removal which provided subsequent reduction in RMSE. The last stage was feature transformation which provided significant reduction in RMSE. The RMSE value for ANN varies significantly after retraining and therefore five iterations at each stage were conducted and the average of these five iterations is recorded in the Table 4.

The results (Table 4) indicate the lowest RMSE value was provided by GPR with exponential kernel. The results of SVR with linear kernel and multiple linear regression were consistently low which was due to non-linear nature of the problem set. The three non-linear methods provided lowest RMSE out of which GPR had a value 0.1542, quadratic SVR had a value of 0.1696 and ANN had an average value of 0.1674. The three regression algorithms with lowest RMSE were evaluated to choose the final model.

The evaluation to select final model was performed by randomly splitting the data into 20% (28 samples) for testing and 80% (113 samples) for training. This splitting was performed so that each model gets the same data for training and testing. The SVR with quadratic kernel was tuned by optimizing its hyper-parameters. Fig. 3 provides the minimum objective vs the number of function evaluations and the selected hyper-parameters for estimated objective function value of 0.093389 was provided in Table 5.

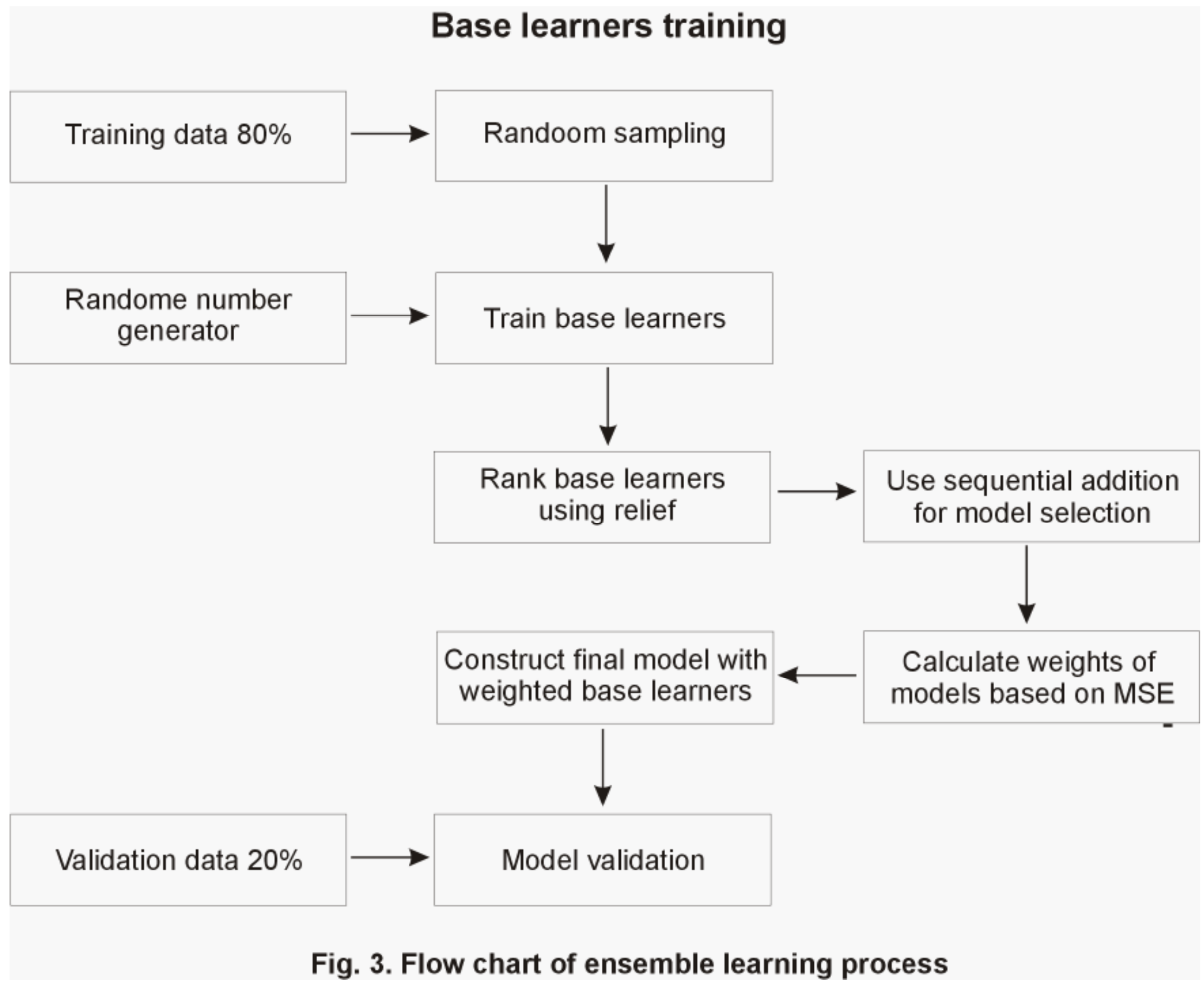

**Fig. 3. Flow chart of ensemble learning process**

**Table 5. Optimized hyper-parameters for quadratic SVR**

| Box constraint | Kernel scale | Epsilon |
|---|---|---|
| 5.375 | 8.5463 | 0.093389 |

The GPR with squared exponential kernel is tuned by optimizing its hyper-parameters. Fig. 4 provided the minimum objective vs the number of function evaluations for GPR and the selected hyper-parameter value for sigma was 0.16517 at estimated objective function value of 0.034994. Figure 5 provides objective function model with 95% confidence interval.

The GPR with squared Exponential kernel is tuned by optimizing its hyper-parameters. Fig. 4 provided the minimum objective vs the number of function evaluations for GPR and the selected hyper-parameter value for sigma was 0.16517 at estimated objective function value of 0.034994. Fig. 5 provided the estimated and minimum objective for 30 iterations whereas Fig. 6 provided the objective function model with 95% confidence interval.

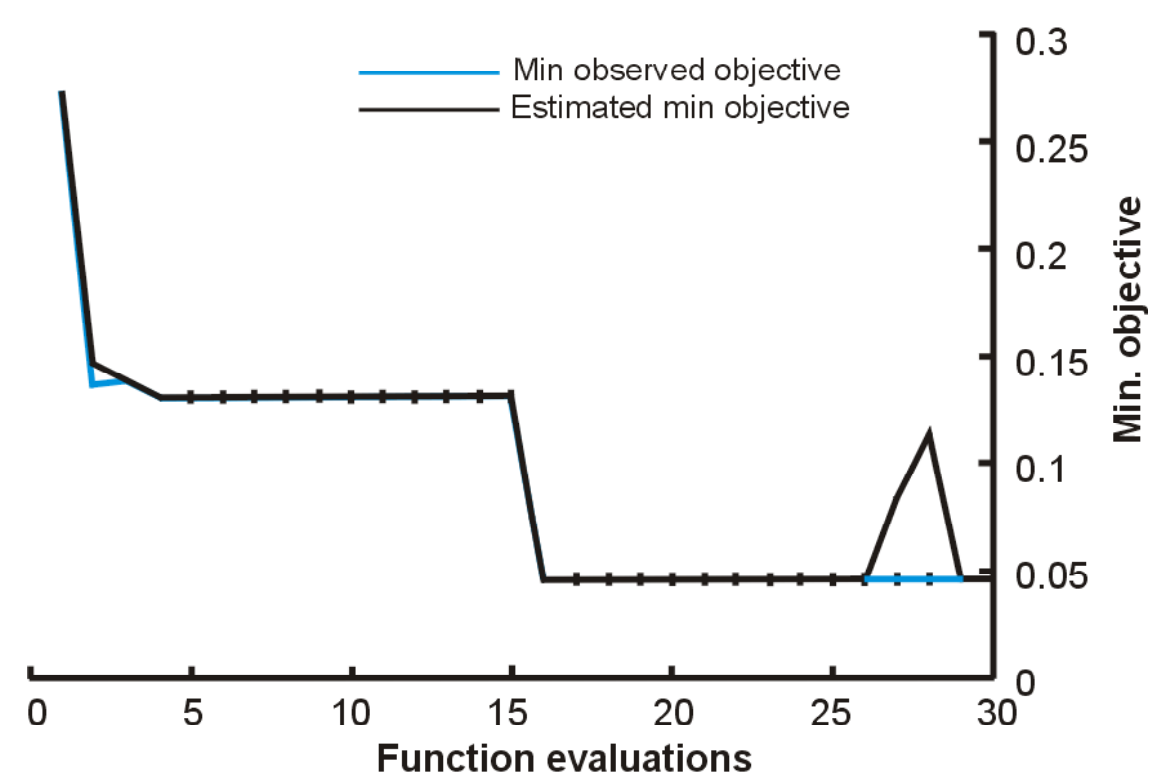

**Fig. 4. Minimum objective vs the number of function evaluations for SVR with quadratic kernel**

Table 6 provided the performance parameters for the three algorithms. The performance is compared based on four evaluation measures among which are MAE, MSE, RMSE and coefficient of determination R2. ANN gave best R2 but it was not stable and the values of MSE and R2 varied a lot with several iteration having same training and test data and different initial conditions. The variance in ANN model was reduced by using ensemble of neural networks. GPR and SVR on the other hand were stable as their performance only vary slightly with change in training and test data.

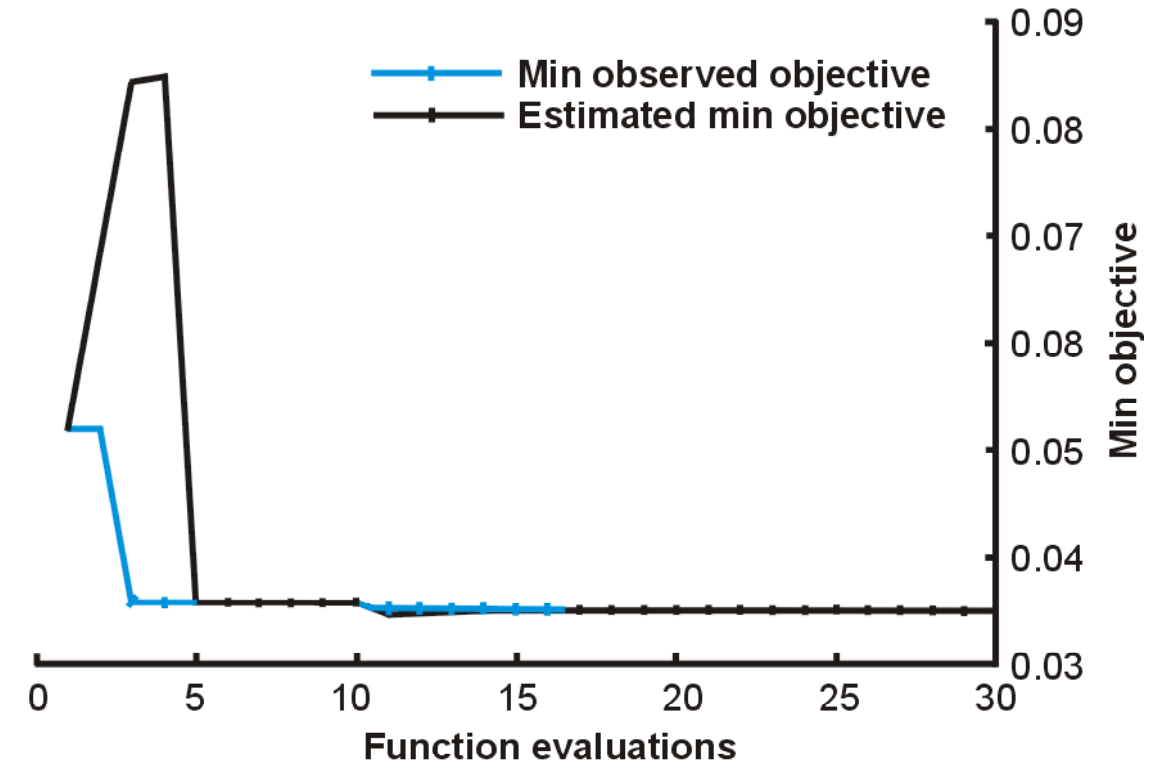

**Fig. 5. Minimum objective vs the number of function evaluations GPR**

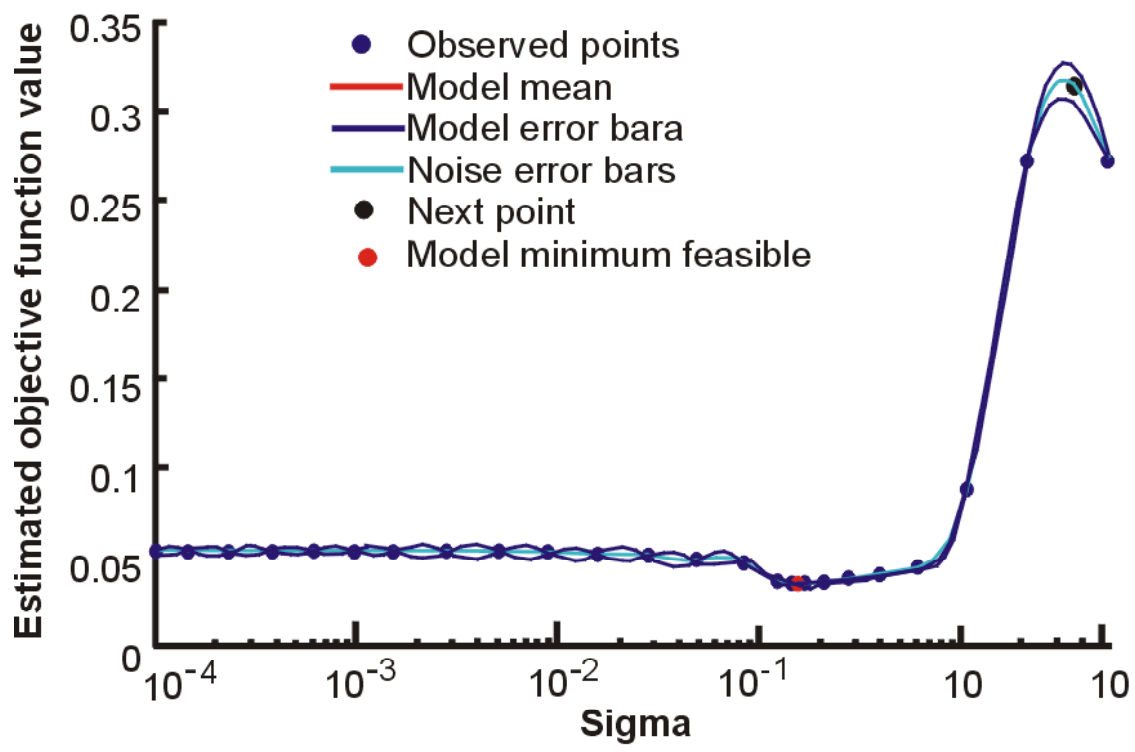

**Fig. 6. Estimated values of objective function for different values of sigma**

The GPR provided best single learner performance whereas the ensembling of neural networks has beaten its performance and provided even higher performance. The proposed model provided highest value of coefficient of determination indicating that the proposed approach has best prediction performance in comparison to baseline approaches. The proposed approach provided a MAE of 0.0942, MSE of 0.0145, RMSE of 0.1204 and R-squared of 0.9461. Similar results were reported by Ahmad *et al.* (20121) and Ahmed *et al.* (2021).

The study highlight that neural networks ensemble have shown best generalization performance. ANN are known to model the complex relationships between independent and dependent variables and their variance can be easily managed by ensemble learning which provide stable models. It is also highlighted that selection of some good performing neural networks is better than adding all the base learners. The selected base learners can be added linearly, through weighted averaging or convex function approach. Moreover, the weights assigned to base learners using ReliefF algorithm were also used to calculate the final weighting function but weight assignment using proposed approach has provided best performance.

**Table 6. Performance parameters of three models**

| Paramerters | SVR | GPR | ANN | RF | Proposed |
|---|---|---|---|---|---|
| MAE | 0.1763 | 0.1328 | 0.1795 | 0.1426 | 0.0942 |
| MSE | 0.0039 | 0.0034 | 0.0300 | 0.0353 | 0.0145 |
| RMSE | 0.0626 | 0.0581 | 0.1732 | 0.1705 | 0.1204 |
| $R^2$ | 0.8357 | 0.9078 | 0.9213 | 0.8942 | 0.9461 |

## CONCLUSION

The present study demonstrated the use of different methods for crop yield forecasting of tea plant. Linear regression models were simplest regression approach as they had low variance but the parametric analysis indicated that predictor and response relationship was not linear and a non-linear approach would be more suitable for the problem. To solve the problem data pre-processing methods were extensively explored and described 32. Five regression algorithms were applied to fit the model among which two were multiple linear regression and support vector machines regression with linear kernel were the two linear algorithms. Support vector machines regression with quadratic kernel, Gaussian process regression with exponential kernel and artificial neural network are the non-linear algorithms. RMSE with 10-fold cross-validation of linear algorithms was quite high so the non-linear algorithms were carried on for further analysis. Fine tuning of these algorithms was performed and their performance was measured on four evaluation metrics. ANN provided the highest values for MAE, MSE and RMSE indicating its poor performance but it provided best result for coefficient of determination (R2). On the other hand, GPR provided lowest values for MAE, MSE and RMSE and the value of R2 is slightly lower than ANN. Moreover, the value of ANN varies a lot with retraining of the model with same training and testing data showing instability so the proposed boosting method utilizing the ANN as weak learner provided relatively stable performance with MAE of 0.0942, MSE of 0.0145, RMSE of 0.1204 and R2 0.9461. The evaluation was performed using 30% hold-out data which was not used in training or validation and it indicate good generalization indicating that the model could be successfully deployed for yield forecasting of tea or other high value crops.

## CONTRIBUTION OF AUTHORS

| Sr. No. | Author's name | Contribution | Signature |
|---|---|---|---|
| 1. | Nisar Ahmed | Conduct the research work and wrote the manuscript | |
| 2. | Hafiz Muhammad Shahzad Asif | Supervised the research work | |
| 3. | Gulshan Saleem | Helped in the research work | |
| 4. | Muhammad Usman Younus | Helped in the research work, prepared and proof read the manuscript | |